\documentclass[10pt, journal,compsoc]{IEEEtran}
\usepackage{amsmath,amsfonts}
\usepackage[ruled, lined, vlined]{algorithm2e}
\usepackage{array}
\usepackage{textcomp}
\usepackage{stfloats}
\usepackage{url}
\usepackage{verbatim}
\usepackage{graphicx}
\usepackage{cite}
\usepackage{subfiles}
\usepackage{amsthm}
\usepackage{booktabs}
\usepackage{subfigure}
\usepackage{balance}

\theoremstyle{definition}
\newtheorem{definition}{Definition}[]

\newcommand{\paratitle}[1]{\vspace{1ex}\noindent \textbf{#1}}

\usepackage[dvipsnames]{xcolor}

\begin{document}

\title{Semantic-Enhanced Representation Learning for Road Networks with Temporal Dynamics}

\author{Yile Chen, Xiucheng Li, Gao Cong, Zhifeng Bao, Cheng Long
\IEEEcompsocitemizethanks{
\IEEEcompsocthanksitem Yile Chen, Gao Cong and Cheng Long are with School of Computer Science and Engineering, Nanyang Technological University, Singapore.
\protect\\
E-mail: yile001@e.ntu.edu.sg, 
\{gaocong, c.long\}@ntu.edu.sg;
\IEEEcompsocthanksitem Xiucheng Li is with School of Computer Science and Technology, Harbin Institute of Technology (Shenzhen), Guangdong, China.
\protect\\
E-mail: lixiucheng@hit.edu.cn;
\IEEEcompsocthanksitem Zhifeng Bao is with School of Computing Technologies, RMIT University, Melbourne, Victoria, Australia.
\protect\\
E-mail: zhifeng.bao@rmit.edu.au;
}
\thanks{Manuscript submitted on xxx, xxx}}

\markboth{Journal of \LaTeX\ Class Files,~Vol.~14, No.~8, August~2021}%
{Shell \MakeLowercase{\textit{et al.}}: A Sample Article Using IEEEtran.cls for IEEE Journals}

\IEEEpubid{0000--0000/00\$00.00~\copyright~2021 IEEE}

\IEEEtitleabstractindextext{
\begin{abstract}
In this study, we introduce a novel framework called Toast for learning general-purpose representations of road networks, along with its advanced counterpart DyToast, designed to enhance the integration of temporal dynamics to boost the performance of various time-sensitive downstream tasks. Specifically, we propose to encode two pivotal semantic characteristics intrinsic to road networks: traffic patterns and traveling semantics. To achieve this, we refine the skip-gram module by incorporating auxiliary objectives aimed at predicting the traffic context associated with a target road segment. 
Moreover, we leverage trajectory data and design pre-training strategies based on Transformer to distill traveling semantics on road networks. 
DyToast further augments this framework by employing unified trigonometric functions characterized by their beneficial properties, enabling the capture of temporal evolution and dynamic nature of road networks more effectively. With these proposed techniques, we can obtain representations that encode multi-faceted aspects of knowledge within road networks, applicable across both road segment-based applications and trajectory-based applications. Extensive experiments on two real-world datasets across three tasks demonstrate that our proposed framework consistently outperforms the state-of-the-art baselines by a significant margin.
\end{abstract}

}

\maketitle

\section{Introduction}\label{sec:intro}


\IEEEPARstart{R}{oad} networks, as a fundamental yet indispensable component in transportation systems, play a crucial role in various downstream transport planning tasks. These tasks include trajectory-based tasks like route inference~\cite{ICDE_route, TKDE22_route} and road segment-based tasks like traffic forecasting~\cite{TKDE_traffic, ICDE19_trafficinfer}.
Recent studies have increasingly focused on deriving effective representations that can capture the intrinsic characteristics of road networks. Such general-purpose representations have the potential to significantly enhance the effectiveness of these varied tasks. 
Given that road networks are essentially a graph, a natural question to ask is whether we can apply graph representation learning models to achieve this goal. Unfortunately, the application of such models is non-trivial due to two issues. 

\begin{figure}[tbp]
    \centering
    \includegraphics[width=0.85\linewidth]{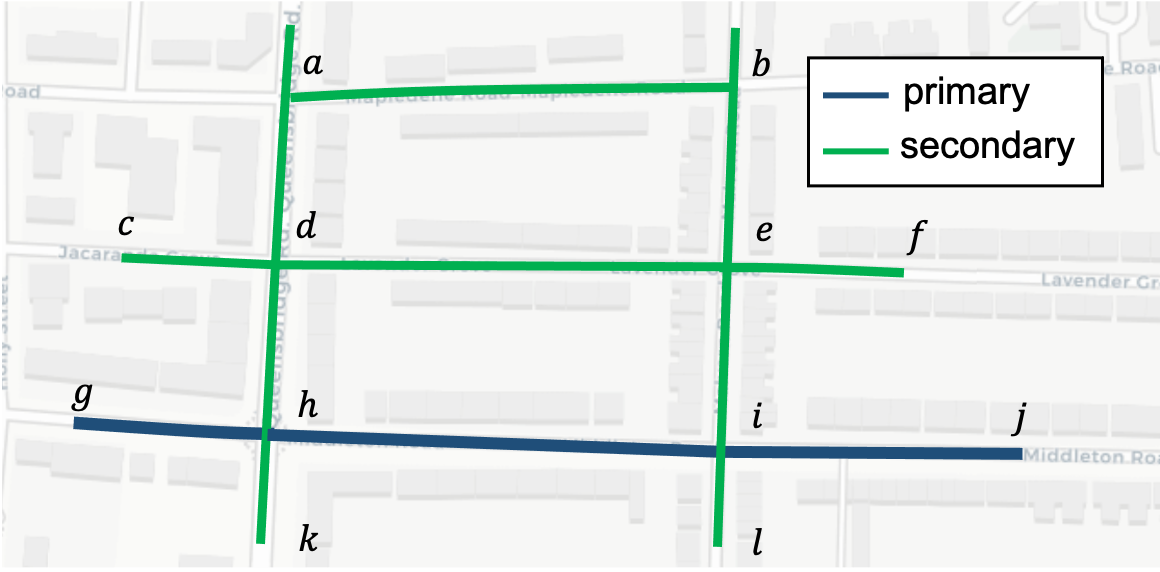}
    \caption{Road network example. Blue line denotes primary roads and green lines denote secondary roads.}
    \label{fig:example}
    \vspace{-3mm}
\end{figure}

The first is the \emph{discrepancies} with regard to the assumptions applied to common graphs and road networks. Most previous graph representation learning methods predominantly target citation graphs~\cite{WWW22_homophily, CIKM23_homophily2} or social networks~\cite{Deepwalk, node2vec}, devising techniques grounded in certain well-established assumptions specific to these types of graphs. 
These assumptions, however, may not be applicable or valid for road networks. For example, a citation graph often exhibits network homophily, where interconnected nodes are more similar than distant nodes. However, this principle does not necessarily translate to road networks, where spatially adjacent road segments might not necessarily exhibit similar traffic patterns. In Fig.~\ref{fig:example}, road segments $dh$, $gh$, $hi$, $hk$ are connected to each other, but primary roads typically have different \emph{traffic patterns}, such as traffic volume, compared to secondary roads since primary roads are travelled more frequently. 

The second is the \emph{feature uniformity} issue inherent in road networks. Features such as road type and lane number are often shared across spatially adjacent road segments. This characteristic is more evident in urban sub-regions with distinct functionalities, such as commercial areas and residential areas, where a large fraction of road networks have the same features within these areas. Such uniformity in road networks can dampen the performance of standard graph representation learning methods, especially graph neural networks (GNN)~\cite{GNN_Survey}. An example of this, illustrated in Fig. \ref{fig:example}, shows that all road segments connected to the target road segment $de$ possess the same features (road type). Similar scenarios are also observed for road segments $cd$, $ad$, $ab$, etc. In such instances, GNN aggregation process renders these road segments indistinguishable when they present the same feature input, leading to a case similar to over-smoothing issues as discussed in~\cite{GNN_oversmooth}. 

Notably, the two issues, \emph{discrepancies} and \emph{feature uniformity}, represent different aspects of potential complications that can coexist on road networks. For example, while the road segment $de$ shares features with its neighbors, it also experiences higher traffic volumes compared to segments $cd$, $ad$, $ab$ due to its positioning on a direct route (path $[c,d,e,f]$), as opposed to a detour (path $[c,d,a,b,e,f]$).


While recent studies have adapted graph representation learning to road network setting, they still have limitations in addressing the two issues. Methods in~\cite{Bigdata18_roadrep, SIGSPATIAL19_roadrep, TIST21} aim to produce representations for road segments and intersections through multi-task learning. They propose to capture the topological structure of road networks while integrating additional classification objectives, such as identifying common attributes between two segments or intersections (e.g., same-way road or stop sign presence). However, these methods heavily rely on the homophily assumption, and thus cannot fully address the first issue. On the other hand, methods in \cite{SIGSPATIAL_roadrep, RoadGNN_KDD20, TITS22_roadrep} adapt GNN to road networks for learning road segment representations. However, they particularly suffer from the second issue in areas with uniform road features. Furthermore, these models focus on capturing certain aspects of road network characteristics, such as topological structure, and consequently, they fall short in learning effective representations that contain multifaceted knowledge about road networks.


We argue that deriving effective road network representations with general applicability requires capturing two types of semantic characteristics: namely \emph{traffic patterns} and \emph{traveling semantics}, in order to address the identified issues. 
Traffic patterns, encompassing factors like traffic volumes, serve as important indicators to enrich the knowledge beyond topological structure, thus overcoming the limitations posed by assumptions for common graphs. Meanwhile, traveling semantics, such as transition patterns, assist in distinguishing road segments that exhibit similar features. As illustrated in Fig. \ref{fig:example}, transition patterns can reveal that the path $[c,d,e,f]$ is more frequently traveled compared to the detour path $[c,d,a,b,e,f]$ between segments $c$ and $f$, thus highlighting the dependencies among road segments. These two types of characteristics represent the most fundamental aspects of road networks. Therefore, their proper encoding and integration are crucial for enriching multi-faced knowledge of road networks, desired in downstream applications.

To this end, we propose a framework called \textsf{Toast} 
to learn general-purpose representations of road networks that can capture both traffic patterns and traveling semantics with dedicated modules. Different from previous models that focused on encoding the graph structure based on skip-gram training objective~\cite{SIGSPATIAL_roadrep, RoadGNN_KDD20}, our method further enables this module to gain awareness of traffic patterns by incorporating an auxiliary traffic context prediction objective. By doing this, the module not only encodes the graph structure of road networks, but also distinguishes connected road segments in terms of traffic patterns, thereby addressing the \emph{discrepancies} issue. 
To tackle the \emph{feature uniformity} issue, we propose to leverage trajectory data to extract traveling semantics for indistinguishable road network fractions caused by uniform features. 
As has been shown in previous studies~\cite{IJCAI17_traj, TIST20}, trajectory data is a rich source of traveling semantics, which can enhance the capture of correlations for road segments.
Inspired by the success of Transformer-based pre-trained models~\cite{BERT}, we employ this architecture to capture transition patterns from trajectory data into representations. Considering the inadequacy of conventional training tasks for text modeling in road network contexts, we design two novel training tasks, \emph{route recovery} and \emph{trajectory discrimination}, tailored to effectively encode the traveling semantics. 
Our framework unifies these two modules, allowing them to focus on encoding complementary aspects of road network characteristics. Both modules are based on self-supervised training paradigms where traffic patterns and traveling semantics are directly treated as training objectives without the need for additional task-specific labels. This ensures that the derived representations are versatile and effective in a range of downstream applications.
Moreover, apart from learning representations of road segments, \textsf{Toast} offers additional advantage of obtaining trajectory representations from the trajectory-enhanced Transformer module. Such capability further enhances the utility for trajectory-based tasks, such as travel time estimation and destination prediction.

\textsf{Toast} has been presented in our prior work~\cite{Toast}, and it
has triggered the development of several subsequent methods for road network representation learning~\cite{EDBT23_roadrep, CIKM22_roadrep,TKDD23_roadrep, PAKDD23_roadrep} to overcome the previously outlined issues. Specifically, some methods propose to refine the conventional methods by incorporating contrastive learning techniques tailored for road networks, such as spatial-aware sampling~\cite{EDBT23_roadrep} and multi-view contrasts between road segments and trajectories~\cite{CIKM22_roadrep}. Others adapt GNN to tackle feature uniformity by hypergraph construction~\cite{TKDD23_roadrep} or transition pattern integration ~\cite{PAKDD23_roadrep}. These methods have achieved encouraging results. However, they, along with other existing studies, aim to learn static representations for road networks. In practice, numerous road network tasks are inherently dynamic: traffic speed on road segments varies over time, and travel times for the same route can differ significantly across different time frames. In this case, there is a growing need to develop time-sensitive road network representations that not only embody better effectiveness but are also more readily applicable to these dynamic tasks.

To develop effective time-sensitive representations, we propose \textsf{DyToast}, an improved version of \textsf{Toast}, equipped with a unified temporal encoding technique that requires minimum model modifications to the original method. \textsf{DyToast} is designed to fuse temporal dynamics into representations by employing learnable trigonometric functions, which exhibit beneficial theoretical properties in road network contexts, into each module. First, apart from refining the original skip-gram objective with traffic patterns in \textsf{Toast}, we augment this module by supplementing the road network graph with transition frequencies for each time frame. This enhancement is complemented by the adoption of parameterization for temporal variations to adeptly capture the evolving patterns for a target road segment in relation to its surrounding road segments. Second, we address the challenge of modeling complex temporal correlations in trajectories with irregular time gaps between consecutive road segments. Traditional absolute or relative positional embeddings in Transformer are insufficient for modeling such irregularities. To resolve this, \textsf{DyToast} integrates the trigonometric function seamlessly into the self-attention mechanism, thereby effectively capturing such irregular and continuous properties. 
Through the proposed temporal encoding technique, \textsf{DyToast} stands out as a solution for capturing not only the dynamic evolution of road segments in relation to their surrounding environment but also the nuanced, higher-order dependencies inherent in trajectories with irregular time intervals. 

To summarize, our contributions are as follows:
\begin{itemize}
    \item We propose a road network representation learning method named \textsf{Toast}, which features with two modules: a traffic context-aware skip-gram module and a trajectory-enhanced Transformer module. These modules are effective in capturing traffic patterns and traveling semantics within road networks. \textsf{Toast} enables the learning of general-purpose representations for road networks, which are beneficial for both road segment-based and trajectory-based applications.
    \item Building upon \textsf{Toast}, we develop an enhanced version, \textsf{DyToast}, which incorporates the ability to capture temporal dynamics. This is achieved through an innovative integration of learnable trigonometric functions, which align seamlessly with \textsf{Toast}.  \textsf{DyToast} excels at encoding temporally nuanced knowledge in both road network and trajectory contexts, offering a more dynamic understanding of road network patterns.
    \item We conduct extensive experiments on three time-sensitive applications on road networks. 
    The results show that \textsf{Toast} performs comparable to existing methods. Furthermore, 
    with the integration of the proposed temporal encoding technique, \textsf{DyToast} demonstrates a significant performance improvement, consistently outperforming baseline methods by a substantial margin. Among these contributions, the first is covered in our preliminary version~\cite{Toast} and the remaining two are newly presented in this extended version.
\end{itemize}
\section{Related Work}
\label{relatedwork}

\subsection{Representation Learning for Road Networks}
Road networks serve as critical components in various intelligent transportation applications, such as traffic inference and forecasting~\cite{ICDE19_trafficinfer, TKDE_traffic}, road attribute prediction~\cite{SIGSPATIAL21_attribute}, and travel time estimation~\cite{VLDB22_traveltime}. In these applications, road network representations are implicitly learned with supervision signals specific to the task at hand. To achieve more generic applicability, recent efforts have also focused on adapting graph representation learning techniques to road networks, aiming to derive general-purpose representations that can benefit a range of tasks. 

Specifically, some studies employ random walk strategies based on the principles of classical Deepwalk~\cite{Deepwalk} and node2vec~\cite{node2vec}. They directly apply the method~\cite{Bigdata18_roadrep}, or modify them to include geo-locality and geo-shape information through multi-task learning~\cite{SIGSPATIAL_roadrep, TIST21}. Another line of research adapts GNN~\cite{GNN_Survey} to road networks. For example, RFN~\cite{SIGSPATIAL19_roadrep, TITS22_roadrep} perform extends GNN to perform multi-view relational fusion by aggregating information at both road segment and intersection levels. HRNR~\cite{RoadGNN_KDD20} adopts a hierarchical GNN approach to model the bottom-up structure of road networks. To address the issues of these methods discussed in Section \ref{sec:intro}, \textsf{Toast} proposes to further capture the knowledge of traffic patterns and traveling semantics by integrating a traffic context prediction objective and pioneering trajectory pre-training strategies. Following \textsf{Toast}, subsequent studies introduce various methods to model such essential knowledge. HyperRoad~\cite{TKDD23_roadrep} implements GNN with hyperedge and hypergraph-based training objectives on hypergraphs constructed from road networks.  TrajRNE~\cite{PAKDD23_roadrep} leverages trajectory data to generate random walks and derive adjacency matrix for GNN, thus combing the studies from two technical branches. JCLRNT~\cite{CIKM22_roadrep} and SARN~\cite{EDBT23_roadrep} further augment \textsf{Toast} with contrastive learning techniques. However, existing methods do not adequately address the dynamic aspect of road network representations, thus resulting in sub-optimal performance for time-sensitive downstream applications. This gap underscores the need for further development in dynamic road network representations, as achieved in \textsf{DyToast}.  

%

\subsection{Trajectory Analysis and Modeling}
Trajectories, representing the movement of vehicles within a city, are a crucial data source to provide supplementary insights for tasks related to road networks~\cite{traj_survey}. In particular, road networks explicitly impose structural constraints that govern the traversal of trajectories, forming the foundations for applications such as route planning~\cite{ICDE_route, TKDE22_route}, anomaly detection~\cite{ICDE20_trajanomaly} and destination prediction~\cite{TKDE20_destination}. Conversely, trajectories provide rich knowledge of traveling semantics for road networks~\cite{IJCAI17_traj}, which can effectively enhance tasks that may not necessarily involve trajectory data. 
For instance, in traffic flow prediction~\cite{KDD21_trajflow} and speed prediction~\cite{AAAI21_trajflow}, trajectories are employed to guide GNN aggregation processes. They are also utilized to extract transition features for region functionality modeling~\cite{IJCAI22_region, TKDE23_regioon}. In the topic of road network representation learning, \textsf{Toast} pioneers the integration of trajectories through pre-training strategies, and therefore produce representations for both road segments and trajectories that are applicable across diverse downstream applications. Following the path of \textsf{Toast}, JCLRNT~\cite{CIKM22_roadrep} adopts a similar model architecture and applies contrastive learning techniques to also obtain both road segment and trajectory representations. In addition, TrajRNE~\cite{PAKDD23_roadrep} utilizes trajectories to derive the adjacency matrix for GNN for modeling higher-order road segment correlations. However, these methods only exploit the sequential aspect in trajectories while neglecting the temporal dimension. This leads to the limitations of not encoding detailed temporal dependencies in the resultant representations, which are important in time-sensitive applications. These limitations are mitigated in \textsf{DyToast}.
\section{Problem Formulation and Overview}
In this section, we present required definitions in this paper, followed by articulating the problem statement. Next, we describe an overview of our proposed framework.  

\subsection{Problem Definitions}
\begin{definition}{(\textbf{Road Networks}).}
Road networks are represented as a directed graph $G=(\mathcal{V},\mathcal{E}, \mathcal{C_{V}})$. $\mathcal{V}$ is a set of vertices, with each vertex $v$ representing a road segment. $\mathcal{E}$ is a set of edges, where each edge $e_{uv} \in \mathcal{E}$ represents a link connecting road segments $u$ and $v$. $\mathcal{C_{V}}$ is a set of features associated with road networks.
\end{definition}

\begin{definition}{(\textbf{Trajectory}).}
A trajectory $T$ is a sequence of sampled points $[p_{i}]_{i=1}^{|T|}$ from the underlying route of a moving object, and each point $p_{i}$ corresponds to a coordinate of latitude and longitude.
\end{definition}

\begin{definition}{(\textbf{Route}).}
A route $\textbf{r}= [(r_i, t_i)]_{i=1}^{n}$ is time-ordered sequence consisting of $n$ adjacent road segments within road networks $G$, where $r_{i}\in \mathcal{V}$ represents the $i$-th road segment in the route, and $t_{i}$ represents the visit timestamp for $r_{i}$.
\end{definition}

In our study, given road networks $G$, a trajectory $T$ is first mapped to the road networks to get its underlying route $\textbf{r}$ by a map matching algorithm~\cite{MapMatching}.

\paratitle{Problem Statement.} Given road networks $G=(\mathcal{V},\mathcal{E},\mathcal{C_{V}})$ and a trajectory dataset $\mathcal{D}=\{T^{(i)}\}_{i=1}^{|\mathcal{D}|}$, we aim to 1) learn vector representations $\{\mathbf{u}_{v}^{t}\}_{v\in\mathcal{V}}$ for road segments within the network, where $t$ is the index of specific time frame (e.g., 8am-9am), and 2) derive the representation $\textbf{u}_{\textbf{r}}$ for any specified route $\textbf{r}$ on the road networks.

It is worth noting that our target is to learn generic representations for both road segments and trajectories rather than task-specific models. The obtained representations are versatile and can be easily applied to road various segment-based and trajectory-based downstream tasks.

\subsection{Framework Overview}\label{sec:framework}


\begin{figure*}[tbp]
    \centering
    \includegraphics[width=0.8\linewidth]{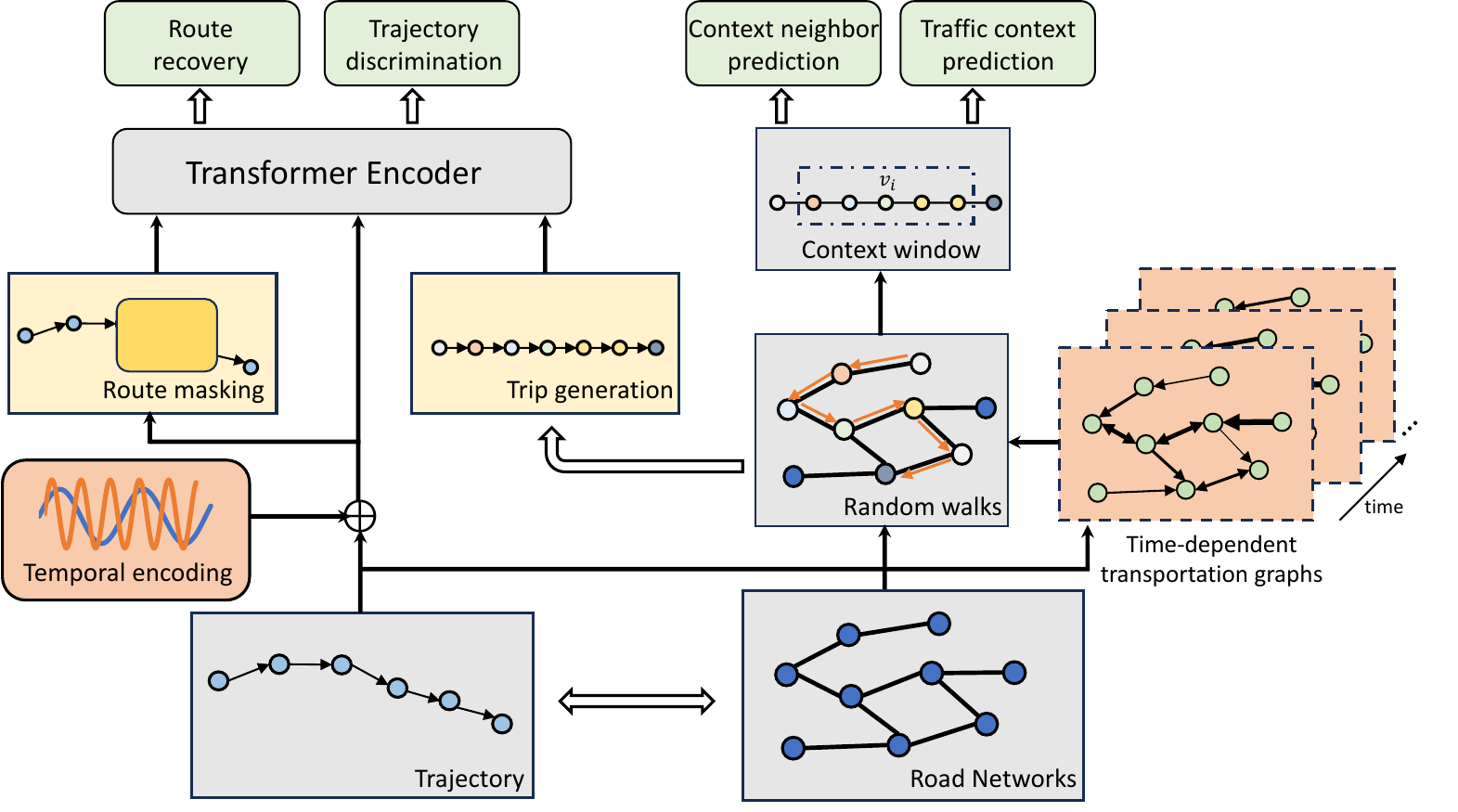}
    \caption{The framework overview of our proposed \textsf{DyToast}.  Components marked in red are designed to integrate temporal dynamics. }
    \label{fig:overview}
    \vspace{-3mm}
\end{figure*}

To obtain general-purpose representations for road networks, we propose \textsf{Toast} to tackle the two issues outlined in Section~\ref{sec:intro} (i.e., discrepancies and feature uniformity). Building upon this approach, we further introduce \textsf{DyToast}, which incorporates the modeling of temporal dynamics prevalent in both road networks and trajectories. An overview of \textsf{DyToast} is presented in Fig.~\ref{fig:overview}. 

To tackle the first issue, we move beyond the conventional graph assumptions typically employed in existing studies~\cite{TIST21}, focusing on mitigating the discrepancies observed in road segments. To achieve this, we extend the skip-gram model~\cite{word2vec}, which is flexible in producing node representations based on a variety of structural assumptions for graphs, to capture traffic patterns (e.g., traffic volume). In addition to the original skip-gram objective of predicting the context neighbors of a target road segment, we introduce auxiliary tasks that predict traffic-related context features (e.g., road category) in a self-supervised manner. Such a multi-task learning paradigm allows the obtained representations to not only encode the graph structure but also differentiate among various traffic patterns that are indicated by these context features.


To tackle the second issue posed by the uniformity of features in various sub-regions, we learn from trajectories to extract traveling semantics on road networks. This includes identifying transition patterns and high-order dependencies between distant regions. To achieve this, we employ Transformer model~\cite{NIPS17_Attn} with two novel pre-training tasks for trajectory data tailored for road network contexts: \emph{route recovery} and \emph{trajectory discrimination}. Specifically, the \emph{route recovery} task involves randomly masking a subsequence of road segments in a given route, and subsequently recovering the masked part based on the remaining segments of the route. The \emph{trajectory discrimination} task aims to discriminate actual routes from the actual trajectories and those generated through random walks on road networks. These proposed techniques within \textsf{DyToast} enable encoding multi-faceted yet mutually enhanced characteristics of road networks into the final representations. Moreover, our framework possesses the capability to produce representations for both individual road segments and trajectories.

Apart from the foundational capabilities conforming to the static characteristics of road networks, we have further augmented it by integrating temporal dynamics into its modules by using unified trigonometric functions. Specifically, we propose to construct a dynamic graph based on transition frequencies at each time frame, and then incorporate trigonometric-based, time-aware functions into skip-gram objective, thus enabling the capture of evolving patterns inherent in these graphs over time. Furthermore, apart from modeling only sequential information in trajectories, we also integrate the function into the self-attention mechanism in trajectory pre-training tasks, allowing the modeling of fine-grained temporal correlations. The modified architecture is adept at handling continuous timestamps with irregular intervals. 
These innovations can substantially improve the effectiveness of produced representations in time-dependent applications.

\section{Methodology}
We elaborate our \textsf{DyToast} framework in this section. We start with preliminaries of the skip-gram model, and then discuss the extended skip-gram model augmented with an auxiliary traffic context prediction objective. Next, we describe the Transformer module and the two trajectory-enhanced pre-training tasks. Finally, we present the temporal encoding techniques integrated into these two modules.

\subsection{Preliminaries: Skip-gram Model}

The skip-gram model was originally introduced in word2vec~\cite{word2vec} to learn embeddings for words. It has been widely adopted in graph representation learning methods later by viewing nodes in a graph as words in a document. This approach involves generating a set of random walks $\mathcal{S}$ on a graph, with each random walk being treated as a sentence. The core objective of the model is to maximize the likelihood of observing the neighborhood nodes within a context window 
given a target node, which equals to minimizing the following loss function:
\begin{equation}
    \mathcal{L}_{SG} = -\sum_{v_{i} \in \mathbf{s}} \sum_{v_{j} \in \mathcal{N}\left(v_{i}\right)} \log p\left(v_{j}|v_{i}\right) 
    \nonumber
\end{equation}

\begin{equation}\label{eq:skip-gram}
 \log p\left(v_{j}|v_{i}\right) = 
    \log \frac{\exp \left(f(v_{i})^{\top} h(v_{j})\right)}{\sum_{v_{j}^{\prime} \in \mathcal{V}} \exp \left(f(v_{i})^{\top} h(v_{j}^{\prime})\right)}   
\end{equation}
where $f, g: \mathbb{N}\rightarrow \mathbb{R}^{d}$ are the embedding functions for target nodes and context nodes respectively, $\mathcal{N}(v_{i})$ is the context neighbors of node $v_{i}$, and $\mathbf{s}$ is a random walk sequence from the set $\mathcal{S}$. 
For computational efficiency, we adopt negative sampling~\cite{word2vec} to optimize the objective in practice, then the objective in Eq.~\ref{eq:skip-gram} can be reformulated as follows:
{\begin{equation}\label{eq:skip-gram-ns}
    \log \left(\sigma\left(f(v_{i})^\top h(v_j)\right)\right) + \sum_{\hat{v}_j \in \mathcal{V}} \log \left(\sigma\left(-f(v_{i})^\top h\left(\hat{v}_j\right)\right)\right)
\end{equation}}
where $\sigma()$ is the sigmoid activation function, and $\mathcal{V}$ is the distribution of the vocabulary for negative sampling. By model training, the final node representations could capture various structural properties (e.g., homophily) via various random walk sampling strategies~\cite{struct2vec, node2vec}.

\subsection{Auxiliary Traffic Context Prediction Objective}

\textsf{Toast} is designed to not only encode 
the structural assumptions of common graphs, but also to incorporate traffic patterns into representations. To achieve this, we propose to extend the skip-gram model by introducing auxiliary traffic context prediction tasks. 
For instance, road segments often have associated traffic context features, such as speed limits and road types, which we treat as auxiliary context information that indicates the traffic patterns of their respective road segments. Based on this, given a target road segment and its context neighbors, our key idea is to first determine the traffic context of the target node, and then to further predict the context neighbors. 
To perform traffic context prediction for a target road segment, we begin by applying binarization to the selected features that are indicative of traffic patterns. For example, if we select road type $c_n$ from the set of traffic context features $\{c_{n}\}_{n=1}^{N}\in \mathcal{C_{V}}$, where $c_n$ has $|c_{n}|$ possible categories, this feature is transformed into a $|c_{n}|$-dimensional label vector where each dimension is 0 or 1, representing the presence of a specific category within the context of the target road segment. 
Formally, given a target road segment $v_{i}$ and its corresponding $N$ types of binarized traffic context features 
$\{c_{n}^{i}\}_{n=1}^{N}$, our goal is to minimize the binary cross entropy loss for any given context feature $c_{n}$:
\begin{equation}
\begin{aligned}
 \mathcal{L}_{c_{n}} = &-\sum_{v_{i} \in \mathbf{s}} \sum_{j=1}^{\left|c_{n}\right|} c_{nj}^{i}\log\sigma(f(v_{i})^{\top} g(c_{nj})) + \\
 &(1-c_{nj}^{i})\cdot\log(1-\sigma(f(v_{i})^{\top}g(c_{nj})))   
\end{aligned}
\end{equation}
where $c_{nj}^{i}$ is the $j$-th entry of the $n$-th binarized feature $c_{n}$ for node $v_{i}$, $f(v_{i})$ is the target embedding for node $v_{i}$, $g(c_{nj})$ is the feature embedding for $c_{nj}$  that is shared across road segments, and $\sigma()$ denotes the sigmoid function.

Then, road segment representations are optimized to produce accurate predictions on both traffic context and context neighbors. This strategy is more appropriate for road network settings than only predicting context neighbors. Moreover, the prediction tasks are structured hierarchically, such that 
the traffic context is utilized to enhance the prediction of context neighbors. In other words, when predicting the context neighbors, instead of only conditioning on the target road segment $v_{i}$ as in Eq.~\ref{eq:skip-gram}, we refine this objective to incorporate traffic context as an additional conditioning factor:
\begin{equation}
\begin{aligned}\label{eq:traffic-skip-gram}
    \mathcal{L}_{SG^{\prime}} &= -\sum_{v_{i} \in \mathbf{s}} \sum_{v_{j} \in \mathcal{N}\left(v_{i}\right)} \log p\left(v_{j}|v_{i},{\boldsymbol{\xi}}(v_{i})\right) \\
    &= -\sum_{v_{i} \in \mathbf{s}} \sum_{v_{j} \in \mathcal{N}\left(v_{i}\right)}\log \frac{\exp \left(\tilde{f}(v_{i})^{\top} \tilde{h}(v_{j})\right)}{\sum_{v_{j}^{\prime} \in \mathcal{V}} \exp \left(\tilde{f}(v_{i})^{\top} \tilde{h}(v_{j}^{\prime})\right)}
\end{aligned}
\end{equation}
where ${\boldsymbol{\xi}}(v_{i}) \stackrel { \mathrm { def. } } { = } [\sigma(f(u_{i})^{\top} g({c}_{nj})]_{j=1, n=1}^{|c_n|, N}$ is the $n$-th predicted traffic context of road segment $v_{i}$. $\tilde{f}(v_{i})$ is the traffic-enhanced target embedding for $v_{i}$, namely, the concatenation of  $f(u_{i})$ and all the predicted traffic context ${\boldsymbol{\xi}}(v_{i})$, and $\tilde{h}(v_{j})$ is the corresponding context embedding for node $v_{j}$. Similarly, we apply the negative sampling strategy as in Eq.~\ref{eq:skip-gram-ns} in practice.

The final objective function is a weighted sum of the modified skip-gram loss and the loss of all auxiliary traffic context prediction tasks. Formally, it is defined as 
\begin{equation}
\mathcal{L} = \mathcal{L}_{SG^{\prime}} + \sum_{n=1}^{N}\delta_{n}\mathcal{L}_{c_{n}}
\end{equation}
where $\delta_{n}$ is the weight of the $n$-th auxiliary task. Compared to the original objective in Eq.~\ref{eq:skip-gram}, we incorporate a broader spectrum of semantic information, particularly traffic patterns, into the representations through our meticulously designed auxiliary tasks. Furthermore, the prediction of context neighbors is also refined with the inclusion of traffic context knowledge.  As a result, this multi-task learning paradigm would produce more robust and effective road network representations. 

\subsection{Transformer and Pre-training Tasks}
To tackle the feature uniformity issue suffered by existing methods, we employ a Transformer model with two novel pre-training tasks specifically designed to extract transition patterns and high-order dependencies on road networks. The effectiveness of Transformer pre-training in modeling text sequences has been extensively validated, particularly in learning semantically rich word representations for numerous downstream tasks~\cite{BERT, GPT3}. Given the sequential nature of trajectory data, we propose to leverage this model to learn representations for road networks. Now we proceed to describe the model details in a bottom-up manner.
\subsubsection{Model Architecture}

\paratitle{Input Embedding Layer.} The road segment representations obtained from the first module serve as the input embeddings in Transformer. 
To preserve the order information in trajectories, learnable positional embeddings are integrated into the input representations as follows:
\begin{equation}\label{eq:input}
    \mathbf{x}_{i} = \mathbf{u}_{i}+ \mathbf{p}_{i}
\end{equation}
where $\mathbf{u}_{i}$ and $\mathbf{p}_{i}$ are road segment representation and positional embedding for the $i$-th road segment, respectively.


\paratitle{Multi-head Self-attention.}
Self-attention mechanism allows the model to selectively focus on correlated parts of the input sequence. We follow the scaled inner-product form of self-attention, which can be described as mapping the representations
of the input sequence to output representations~\cite{NIPS17_Attn}. Formally, it is defined as
\begin{equation}
    \operatorname{Attention}(\mathbf{Q}, \mathbf{K}, \mathbf{V})=\operatorname{softmax}\left(\frac{\mathbf{Q} \mathbf{K}^{\top}}{\sqrt{d_{k}}}\right) \mathbf{V}
\end{equation}
where $\mathbf{Q}$, $\mathbf{K}$ and $\mathbf{V}$ are the query, key, and value matrix respectively derived from a linear projection on the representations
of the input trajectory, and $d_k$ is the vector dimension, which is set to be the same for all the three matrices.

In our work, we adopt multi-head self-attention to model trajectory sequences. Specifically, the representations of input trajectory are projected into $h$ sets of different queries, keys, and values to perform the self-attention mechanism, which has been shown to achieve better performance. Given the input representations $\mathbf{X} = [\mathbf{x}_{1},...,\mathbf{x}_{N}]\in \mathbb{R}^{N\times d_\mathrm{in}}$ with length $N$ where $\mathbf{x}_{i}$ is the representation of the $i$-th road segment in the trajectory after the embedding encoding layer, the output representations $\mathbf{Z}$ $=$ $ [\mathbf{z}_{i}, \mathbf{z}_{2},..., \mathbf{z}_{N}]\in \mathbb{R}^{M \times d_\mathrm{out}}$ are produced as follows:
\begin{equation}
\begin{aligned}
    & \mathbf{Z} = \operatorname{MH-Attn}(\mathbf{X}) = [\mathrm{head}_{1}, \ldots, \mathrm{head}_{h}] \cdot \mathbf{W}^{O} \\
    & \mathrm{head}_{i} =\operatorname{Attention}\left(\mathbf{X} \mathbf{W}_{i}^{Q}, \mathbf{X} \mathbf{W}_{i}^{K}, \mathbf{X} \mathbf{W}_{i}^{V}\right)
\end{aligned}
\end{equation}
where $\mathbf{W}_{i}^{Q}$, $\mathbf{W}_{i}^{K}$, $\mathbf{W}_{i}^{V}$$\in\mathbb{R}^{d_\mathrm{in} \times d_\mathrm{in}/h}$, $\mathbf{W}^{O}$$\in\mathbb{R}^{d_\mathrm{in} \times d_\mathrm{out}}$ are self-attention parameters.

\paratitle{Position-wise Feed-forward Network.}
After multi-head self-attention component, the output representations $\mathbf{Z}$ are sent to a fully connected feed-forward network as follows:
\begin{equation}
\begin{aligned}
    \operatorname{FFN}(\mathbf{Z}) 
    &= \Phi\left(\mathbf{Z} \mathbf{W}_{1}+\mathbf{b}_{1}\right) \mathbf{W}_{2}+\mathbf{b}_{2}    
\end{aligned}
\end{equation}
where $\Phi()$ is the ReLU activation function, $\mathbf{W}_{1}$, $\mathbf{W}_{2}$, $\mathbf{b}_{1}$ and $\mathbf{b}_{2}$ are parameters of the feed-forward network.

\paratitle{Model Stacking.}
It is usually beneficial to learn more complex transition patterns in trajectory data by stacking multiple layers. In particular, each layer is composed of two sub-layers, namely multi-head self-attention and position-wise feed forward network,  connected by residual connection and layer normalization as follows: 
\begin{equation}
\begin{aligned}
    \mathbf{Z^{\prime}} &= \operatorname{LayerNorm}( \mathbf{X}+\operatorname{MH-Attn}(\mathbf{X})) \\
    \mathbf{X^{\prime}} &= \operatorname{LayerNorm}( \mathbf{Z^{\prime}}+\operatorname{FFN}(\mathbf{Z^{\prime}})) \\
\end{aligned}
\end{equation}
where $\operatorname{LayerNorm}$ denotes layer normalization and $\mathbf{Z^{\prime}}$ denotes the final output representations which are passed as the input to the subsequent layer of Transformer.

\subsubsection{Model Pre-Training}
\label{subsec:model_learning}
Despite the capabilities of Transformer model, a critical concern is how to ensure that the derived representations adequately encode traveling semantics within road networks. The model's effectiveness largely depends on the design of loss functions that are adequately tailored to the specific domain (e.g., language~\cite{BERT}, image~\cite{ImageBERT}, video~\cite{VideoBERT}). To this end, it is important to devise appropriate pre-training tasks that demand the comprehension of traveling semantics on road networks. 


Common pre-training tasks employed in Transformer models include masked language modeling~\cite{BERT}, next token prediction~\cite{GPT3}, and other tasks such as next sentence prediction~\cite{BERT} and sentence order prediction~\cite{ALBERT}. While these tasks can generate effective text representations, they cannot achieve our target under road network settings. 
For example, consider the masked language modeling task, where each word in a sequence is randomly masked at a certain probability (e.g., 15\%), and the model is then tasked with predicting these masked words. However, this task is less effective when applied to trajectories within road network contexts. This is because two consecutive road segments in a trajectory must be connected in road networks. When a road segment is masked within such a sequence, it can often be trivially inferred from the knowledge of the graph structure and its given adjacent road segments, as t represents the only segment that makes the sequence a valid route. Since the graph structure is well captured by the skip-gram objective, this task may not contribute additional valuable information for the representations. Moreover, in the next token prediction task, the focus is primarily on forward prediction within a sequence, which does not contribute to a comprehensive understanding of the entire trajectory. In addition, sentence-level pre-training tasks do not naturally align with the goals of road network road network representation learning.


To this end, we propose two novel pre-training tasks for trajectory data within road networks: \emph{route recovery} and \emph{trajectory discrimination}. These tasks are tailored to effectively encode the traveling semantics into  representations.

\paratitle{Route Recovery.} 
Different from the masked language model task where every independent word is randomly masked, we mask a continuous sequence of road segments within a trajectory to make it into a partially observed route. In particular, given a route, we randomly mask 40\% of the consecutive road segments in the sequence. This task prevents the trivial recovery of the masked road segments based solely on the awareness of the graph structure. Instead, it requires the representations to capture more complex transition patterns and accurately identify the most likely options for the masked segments. The model is trained by the cross entropy loss between masked road segments and the predicted ones.

\paratitle{Trajectory Discrimination.}
This task is designed to enhance the model's ability to distinguish real trips from generated ones. Real trips are sampled from our trajectory databases, while fake trips are generated through random walks on road networks. We train the model to minimize the prediction error for these two types of trips. The purpose of this task is two-fold. First, it provides an alternative way for the model to capture transition patterns. After training, the model is able to identify fake trips by recognizing sub-sequences that do not follow the normal transition patterns. Second, this task offers a holistic perspective on traveling semantics across road networks. Trajectories naturally span various regions of the network, and by accurately identifying actual trips, especially those that occur frequently between distant regions, the model can effectively capture high-order dependencies and correlations between distant road segments. 

\subsection{Encoding Temporal Dynamics} \label{subsec:temporal}

\subsubsection{Dynamic Extension for Traffic-enhanced Skip-gram}
\paratitle{Time-dependent Transportation Graph Construction.}
To fuse the enhanced skip-gram model with dynamicity, we propose to construct time-dependent transportation graphs. These graphs are designed to characterize not only the static structural information of road networks, as captured in the previous module, but also the dynamic transition information. Specifically, time is split into discrete time frames, each representing a specific period (e.g., 8am-9am). For each time frame $t$, we augment the road network graph $G$ to form $G^{t_{i}}=(\mathcal{V}, \mathcal{E}_{t}, \mathcal{C_{V}})$, which reflects the holistic transportation condition at each time frame $t$. Here, $\mathcal{V}$ and $\mathcal{C_{V}}$ remain as the road segments and their corresponding features, respectively, while each edge $e_{ij}^{t} = (v_i, v_j, w_{ij}^{t}) \in \mathcal{E}_{t}$ denotes an adjacent link between road segments $i$ and $j$ with an associated weight $w_{ij}^{t}$, defined as:
\begin{equation}
    w_{ij}^{t} = \gamma \times e_{ij} + \mathrm{count}^{t}(v_{i} \rightarrow v_{j})
\end{equation}
where $e_{ij} \in \{0,1\}$ indicates the presence of a structural edge on road networks, $\mathrm{count}(v_{i} \rightarrow v_{j})$ denotes the frequencies of transitions between road segment $i$
and $j$ within time frame $t$, and $\gamma$ is the hyperparameter to balance the influence of these two terms. By constructing a collection of transportation graphs, we gain a holistic view of vehicle movements for each time frame, thus supplementing the static traffic context, typically indicated by features such as speed limits and road types, with dynamic traffic patterns.

\paratitle{Trigonometric Parameterization.}
Equipped with the time-dependent transportation graphs, we perform the strategy as in Eq.~\ref{eq:skip-gram-ns} while integrating a novel temporal encoding technique, which modifies the target embedding function $f$ to additionally condition on the time variable $t$:
{\begin{equation}
    \log \left(\sigma\left(f(v_{i}, t)^\top h(v_j)\right)\right)+ \sum_{\hat{v}_j \in \hat{\mathbb{V}}} \log \left(\sigma\left(-f(v_{i}, t)^\top h\left(\hat{v}_j\right)\right)\right)
\end{equation}}
Here, the proximity degree calculated by $f(v_{i}, t)^\top h(v_j)$ is essential to capture the evolving patterns inherent in the transportation graphs over time. To achieve this, we employ sinusoidal function: $\psi(t): \mathbb{R} \rightarrow \mathbb{R}^{d}$ to model road segment representations specific to time frame $t$, defined as:

\begin{equation}\label{eq:sinusoidal}
\psi(t) = [\cos (\mathbf{w}_t \odot t) || \sin (\mathbf{w}_t \odot t)] \in \mathbb{R}^{d},     
\end{equation}
where $\mathbf{w}_{t} \in \mathbb{R}^{d/2}$ is a learnable parameter to control the frequencies, $\odot$ indicates the broadcast multiplication between vector and scalar, and $||$ denotes vector concatenation. Then the proximity degree is expressed as:

\begin{equation}
\begin{aligned}
    f(v_{i}, t)&^{\top}h(v_{j}) = \mathbf{u}_{i}^{t \top}\mathbf{v}_{j} = (\mathbf{u}_i + \psi(t))^{\top} \mathbf{v}_{j} \\
    &= \mathbf{u}_{i}^{\top}\mathbf{v}_{j} + \sum_{k=1}^{{d}/{2}} v_{j,k} \cos(w_{t,k}t) + v_{j,k+\frac{d}{2}} \sin(w_{t,k}t),
\end{aligned}
\end{equation}
This formulation belongs to the trigonometric polynomial $a_0 + \sum_{k=1}^{K} a_k \cos (k x)+ b_k \sin (k x)$ where $a_{0}, \ldots, a_{K}$, $b_{1}, \ldots, b_{K} \in \mathbb{R}$. Such a function, with suitably selected coefficients, can approximate any periodic continuous functions defined over an arbitrarily closed interval~\cite{polynomial, word2fun}. This capacity aligns well with the typical characteristics of road networks, which often exhibit fluctuating and periodic patterns across days that can be learned by the proposed technique. Consequently, this approach significantly enhances the effectiveness of the representations in capturing temporal aspects.  

\subsubsection{Temporal Feature Integration in Transformer}
The original Transformer applied in our previous study is limited to only encoding sequence ordering information through pre-defined or learned positional embeddings~\cite{NIPS17_Attn, relative_position}. To further effectively model temporal correlations encoded in road segments within trajectories, an intuition approach is to employ discrete embeddings to represent specific time intervals, with each embedding associated with a particular time bin (e.g., 30 seconds). However, this approach has significant drawbacks. It becomes challenging to select appropriate intervals to obtain discrete embeddings, particularly when dealing with irregular time intervals between consecutive road segments. In addition, the method also fails to model the fine-grained correlations for two different intervals that fall in the same time bin.

In light of this, we adopt the sinusoidal function as in Eq.~\ref{eq:sinusoidal} to model continuous visit timestamp $t_i$ at road segment $r_{i}$ within Transformer. Specifically, the self-attention mechanism computes dot-product between two encoded temporal features $\psi(t_i)$ and $\psi(t_j)$ as:

\begin{equation}\label{eq:dot_product}
\psi(t_i)\cdot \psi(t_j)= \mathcal{K}(t_i,t_j) = \mathbf{1}^\top \cos (\mathbf{w}_t \odot (t_{i}-t_{j}))
\end{equation}
This means that such a temporal encoding function can be viewed as a translation-invariant kernel (i.e., $\mathcal{K}(x,y) = \mathcal{K}(x+c,y+c)$), which offers several advantages. First, the translation invariant property allows the model to focus on the interval gaps between timestamps, which are more informative for denoting travel time on road segments than the absolute values of timestamps. Second, it enables direct modeling of correlations on continuous timestamps without manually selecting the intervals for discrete embeddings and thus reducing the information loss. Then the input representations into the Transformer are modified as follows:
\begin{equation}\label{eq:embedding}
\hat{\mathbf{x}}_i = \mathbf{x}_i + \psi(t_{i}) + \mathbf{e}_t
\end{equation}
where $\mathbf{x}_i$ is the representation for road segment ${i}$ within a trajectory as in Eq.~\ref{eq:input}, and $\mathbf{e}_t$ is the time frame representation which encodes coarse-grained temporal information corresponding to the start time of the trajectory. To facilitate the modeling of temporal correlations, we initialize the parameter from a normal distribution $\mathbf{w}_t \sim \mathcal{N}(0, \sigma^{-2})$. By doing this, Eq.~\ref{eq:dot_product} approximates the Gaussian kernel (i.e., $\psi(x)\cdot \psi(y)\approx \operatorname{exp}(-\|x-y\|^{2}/\sigma^{2})$) over its temporal differences~\cite{NIPS_kernel}. This introduces a useful inductive bias of $L_2$ distances as the starting point in the model, enhancing its ability to capture temporal dynamics within trajectories.

\subsubsection{Remarks} 
The adoption of the sinusoidal function in our framework offers a unified solution to significantly enhance the temporal dynamics encoding within the proposed two modules. On the one hand, it complements the skip-gram objective by capturing fluctuating and periodic patterns inherent in time-dependent transportation graphs. This capability contributes to a holistic understanding of the entire road network, thereby providing a macroscopic perspective of temporal evolution. On the other hand, this function serves as an effective way of encoding visit timestamps within trajectories. Its seamless integration into the Transformer's self-attention mechanism enables the model to perform continuous and translation-invariant modeling of fine-grained temporal correlations. In this way, it facilitates the capability of understanding microscopic higher-order dependencies for road segments from each individual trajectory. 

As a result, the representations produced by $\textsf{DyToast}$ -- both in terms of road segment representations and trajectory representations derived from Transformer outputs -- are enriched with multi-faceted characteristics enhanced by the inclusion of temporal dynamics. The model's ability to capture dynamic traffic patterns and traveling semantics ensures that these representations are highly effective for time-sensitive downstream applications.


        
\section{Experiments}
In this section, we compare our proposed framework against other methods applied in road network representation learning. We perform extensive experiments on two real-world datasets, and across three time-sensitive tasks to test the effectiveness of the learned representations for both road segment-based and trajectory-based applications.


\subsection{Datasets}
We utilize two datasets comprising road networks and trajectory data from two cities, $\emph{Chengdu}$ and $\emph{Xi'an}$. The road networks are obtained from OpenStreetMap~\cite{OSM}, while the trajectory data was obtained from the ride-hailing company DiDi, spanning the month of November 2016. To verify the effectiveness of leveraging trajectory data in road network representation learning, we filter out the road segments not covered by trajectory data. We further apply map matching algorithm~\cite{MapMatching} to convert the GPS records from trajectories into sequences of road segments. The statistics of the datasets are presented in Table \ref{tab:dataset}. 

\begin{table}[htp]
\centering
\caption{Statistics of the datasets}
\vspace{-2mm}
\resizebox{0.9\linewidth}{!}{
\begin{tabular}{c|c|c|c}
\toprule
Dataset &\#Road Segments & \#Edges & \#Trajectories \\ \midrule
Chengdu & 6.125           & 15,933          & 5,266,120        \\ 
Xi'an   & 5,146           & 12,804          & 2,533,359       \\ \bottomrule
\end{tabular}}
\label{tab:dataset}
\end{table}


\subsection{Compared Methods}
To evaluate the performance of our proposed \textsf{Toast} and \textsf{DyToast}, we conduct extensive comparisons with 9 baseline methods, described as follow.

\noindent\textbf{Conventional graph learning methods:}
\begin{itemize}
    \item \textbf{node2vec}~\cite{node2vec}: It employs biased random walks on road networks to explore neighborhood of each road segment, capturing both local and high-order structural knowledge within road networks.
    \item \textbf{GCN}~\cite{GCN}: It is the implementation of graph convolutional network for road networks. The model is trained to reconstruct the original road network structure. 
    \item \textbf{GAT}~\cite{GAT}: It is the implementation of graph attention network that applies attention mechanism in the aggregation operation. The model is trained to reconstruct the original road network structure.
\end{itemize}
\noindent\textbf{Standard road network representation learning methods:}
\begin{itemize}
    \item \textbf{SRN2Vec}~\cite{TIST21}: It adopts multi-task learning to make road segments with similar properties, such as spatial closeness and road shapes, close in the representation space.
    \item \textbf{HRHR}~\cite{RoadGNN_KDD20}: It employs a hierarchical GNN to model different semantic levels in road networks. It utilizes two reconstruction tasks to learn the inter-relationships between the three layers of the road network hierarchy.
    \item \textbf{RFN}~\cite{TITS22_roadrep}: It utilizes a multi-view GNN to learn representations from both node-relational and edge-relational perspectives of road network graphs.
    \item \textbf{SARN}~\cite{EDBT23_roadrep}: It adapts graph contrastive learning techniques to road networks by integrating spatial proximity and distance-based negative sampling in its data augmentation stage.
\end{itemize}

\noindent\textbf{Trajectory-enriched representation learning methods:}
\begin{itemize}
    \item \textbf{JCLRNT}~\cite{CIKM22_roadrep}: It aims to derive both road segment and trajectory representations through contrastive learning, including the contrast between road-road, road-trajectory and trajectory-trajectory interactions. 
    \item \textbf{TrajRNE}~\cite{PAKDD23_roadrep}: It leverages trajectory data to construct adjacency matrix for GNN aggregation, and incorporates road context information using techniques similar to SRN2Vec.
\end{itemize}

\subsection{Experimental Settings}
\subsubsection{Downstream Tasks}
Given the capability of \textsf{DyToast} to produce both road segment and trajectory representations, we evaluate its effectiveness on three tasks in time-sensitive settings: road traffic inference, travel time estimation, and destination prediction.

\paratitle{Road speed inference.}
Our target is to infer dynamic traffic speeds on all road segments in scenarios where only partial traffic speed observations are available. Specifically, at each time frame, and aim to infer the average traffic speed for road segments that have missing values, using the model trained on traffic speeds from other road segments. For evaluation, we split a day into one-hour time frames, and extract the speed information at each time frame using the aggregated records across the days from the dataset to avoid data sparsity. Then we randomly mask out 20\% of the traffic speed data at each time frame. We apply 5-fold cross validation to evaluate the performance of all the compared methods. In these methods, the learned road segment representations are utilized as input features into a two-layer fully-connected neural network, which functions as a regressor for this task.

\paratitle{Travel time estimation.} Our target is to estimate the travel time of trajectories that start at varying time frames. Specifically, for methods that do not inherently generate trajectory representations, we produce the trajectory representations by employing a two-layer Transformer to process the representations of road segments as inputs. In contrast, for trajectory-enriched methods that are equipped to produce trajectory representations, we exclude timestamps in the inputs to avoid data leakage. Subsequently, the derived or directly produced trajectory representations are fed into a linear layer to get the prediction of travel time for all the methods. We define each time frame as one hour, and use 80\% trajectory data for pre-training tasks when applicable, while the remaining 20\% is further partitioned into 4:1 ratio for the task-specific training and evaluation stages. 

\paratitle{Destination prediction.} Our target is to predict the destination road segment of trajectories that start at varying time frames. Specifically, we utilize the initial 50\% of road segments as partial trajectories to produce their corresponding representations. Then these representations are fed into a linear layer to classify the destination road segment. The strategies for deriving trajectory representations, as well as the data partitioning settings, are consistent with those outlined in travel time estimation task.

For the tasks of road speed inference and travel time estimation, we use mean absolute error (MAE) and root mean square error (RMSE) as evaluation metrics. For the task of destination prediction, we use Top-$N$ accuracy (Acc@$N$) as metrics to evaluate the proportion of instances where the actual destination road segment appears in top-$N$ predictions ranked by highest probabilities.

\begin{table*}[tbp]
\centering
\caption{Performance of the compared methods on the Chengdu dataset.}
\vspace{-2mm}
\resizebox{0.7\linewidth}{!}{
\begin{tabular}{c|cc|cc|cc}
\toprule
Task     & \multicolumn{2}{c|}{Road Speed Inference} & \multicolumn{2}{c|}{Travel Time Estimation} & \multicolumn{2}{c}{Destination Prediction} \\ \midrule
Metric   & MAE                & RMSE                & MAE                 & RMSE                 & Acc@5               & Acc@10               \\ \midrule
node2vec & 9.72               & 12.93               & 98.92               & 142.25               & 0.502               & 0.593                \\
GCN      & 9.13               & 12.24               & 97.68               & 141.14               & 0.494               & 0.582                \\
GAT      & 9.30               & 12.48               & 96.76               & 140.20               & 0.491               & 0.578                \\
SRN2Vec  & 8.66               & 11.75               & 96.33               & 139.51               & 0.554               & 0.644                \\
HNRN     & 8.71               & 11.84               & 93.46               & 136.72               & 0.545               & 0.636                \\
RFN      & 8.95               & 12.16               & 92.58               & 135.66               & 0.477               & 0.562                \\
SARN     & 8.95               & 12.18               & 94.80               & 137.28               & 0.524               & 0.617                \\
TrajRNE  & 9.01               & 12.23               & 93.69               & 136.91               & 0.563               & 0.655                \\
JCLRNT   & \underline{7.93}               & \underline{11.05}               & 87.92               & 128.70               & \underline{0.601}               & \underline{0.694}                \\
Toast    & 8.72               & 11.87               & 8\underline{3.04}               & \underline{119.77}               & 0.587               & 0.679                \\
DyToast  & \textbf{6.96}               & \textbf{10.08}               & \textbf{75.93}               & \textbf{111.48}               & \textbf{0.632}               & \textbf{0.723}      \\
\bottomrule
\end{tabular}}
\label{tab:result_chengdu}
\end{table*}

\begin{table*}[tbp]
\centering
\caption{Performance of the compared methods on the Xi'an dataset.}
\vspace{-2mm}
\resizebox{0.7\linewidth}{!}{
\begin{tabular}{c|cc|cc|cc}
\toprule
Task     & \multicolumn{2}{c|}{Road Speed Inference} & \multicolumn{2}{c|}{Travel Time Estimation} & \multicolumn{2}{c}{Destination Prediction} \\ \midrule
Metric   & MAE                & RMSE                & MAE                 & RMSE                 & Acc@5               & Acc@10               \\ \midrule
node2vec & 8.40               & 10.86               & 123.53               & 190.79               & 0.476               & 0.572                \\
GCN      & 7.83               & 10.21               & 120.76               & 187.24               & 0.438               & 0.530                \\
GAT      & 8.06               & 10.47               & 118.23               & 185.85               & 0.447               & 0.544                \\
SRN2Vec  & 7.46               & 9.88               & 115.84               & 182.28               & 0.522               & 0.613                \\
HNRN     & 7.72               & 10.13               & 114.37               & 180.63               & 0.516               & 0.608                \\
RFN      &  7.91              & 10.34               & 116.08               & 182.01               & 0.467               & 0.561                \\
SARN     & 7.64               & 10.04               & 112.15               & 178.14               & 0.488               & 0.580                \\
TrajRNE  & 7.79               & 10.14               & 115.13               & 181.62               & 0.526               & 0.625                \\
JCLRNT   & \underline{6.94}               & \underline{9.28}               & 108.69               & 172.28               & \underline{0.557}               & \underline{0.662}                \\
Toast    & 7.57               & 9.98               & \underline{108.12}               & \underline{171.26}               & 0.551               & 0.659                \\
DyToast  & \textbf{6.49}               & \textbf{8.92}               & \textbf{92.64}               & \textbf{145.79}               & \textbf{0.590}               & \textbf{0.688}      \\
\bottomrule
\end{tabular}}
\label{tab:result_xian}
\end{table*}

\subsubsection{Parameter Settings}
We set the dimension of the representations for both road segments and trajectories to be 128 for all the compared methods. For our auxiliary traffic context prediction task, we select road type as the prediction objective where the weight is set to be 1. Furthermore, during the Transformer pre-training phase, we employ a mask ratio of 40\%. We set the number of layers in Transformer to be 2 and the head number to be 4. We apply 30 training epochs for both modules iteratively in our experiments. For the baseline methods, we adhere to the default configurations as described in their respective papers.

\subsection{Performance Comparison}
The results of all the methods across the three tasks on the Chengdu and Xi'an dataset are presented in Table~\ref{tab:result_chengdu} and Table~\ref{tab:result_xian}, respectively. Then we have several observations. 

First, methods such as node2vec, GCN and GAT, which are not specifically designed for road networks, yield the worst results among the baselines. This highlights the importance of developing approaches to tackle the distinctive characteristics of road networks. Second, methods like SRN2Vec, HNRN, RFN and SARN, which focus on capturing road-specific features and spatial information, demonstrates improved performance in the road speed inference task compared to those generic graph representation learning methods. However, they are less effective in trajectory-based tasks due to a lack of capability in modeling high-order dependencies among road segments, which are usually reflected in trajectory data. Furthermore, methods that leverage trajectory data for extracting high-order dependencies, including TrajRNE, JCLRNT, and Toast, although not showing further improvement in road speed inference task, exhibit enhanced performance in trajectory-based tasks. Fourth, among all the baselines, Toast achieves the best performance in travel time estimation, due to its tailored sequence modeling for trajectory pre-training. On the other hand, JCLRNT, building upon Toast by integrating contrastive learning objectives, facilitates enriched interactions between road segments and trajectories, thus enhancing the effectiveness for these two data modalities. Therefore, it achieves the superior performance in both road speed inference and destination prediction. Lastly, the baseline methods generally fall short in capturing the dynamic aspects of road network representation learning. In contrast, \textsf{DyToast} introduces a unified temporal encoding strategy, adeptly adapting to the temporal dynamics inherent in road networks. As a result, our proposed $\textsf{DyToast}$ outperforms all the compared methods with substantial margin across these time-sensitive tasks.

\subsection{Model Analysis}
\subsubsection{Ablation Study}
We conduct an ablation study by removing different components to investigate their contributions to the performance. Specifically, we test the effectiveness of components that enrich temporal dynamics, given that other components have been previously evaluated in the Toast study. For this purpose, we compare \textsf{DyToast} with the following variants:
\begin{itemize}
    \item \textbf{DyToast-G}: it removes the construction of time-dependent transportation graphs, relying instead on the static road network structure while still applying the proposed techniques for encoding temporal dynamics.
    \item \textbf{DyToast-S}: it removes the temporal encoding mechanism within the traffic-enhanced skip-gram module.
    \item \textbf{DyToast-T}: it removes the temporal feature integration mechanism within the Transformer module, omitting the capture of fine-grained temporal correlations in trajectory data.
    \item \textbf{DyToast-ST}: it is a combination of the variants of DyToast-S and DyToast-T by removing both components in skip-gram and Transformer modules. 
\end{itemize}

\begin{figure}[t]
  \centering
  \subfigure[Road speed inference] {
        \begin{minipage}[t]{1.0\linewidth}
        \centering
        \includegraphics[width=0.95\linewidth]{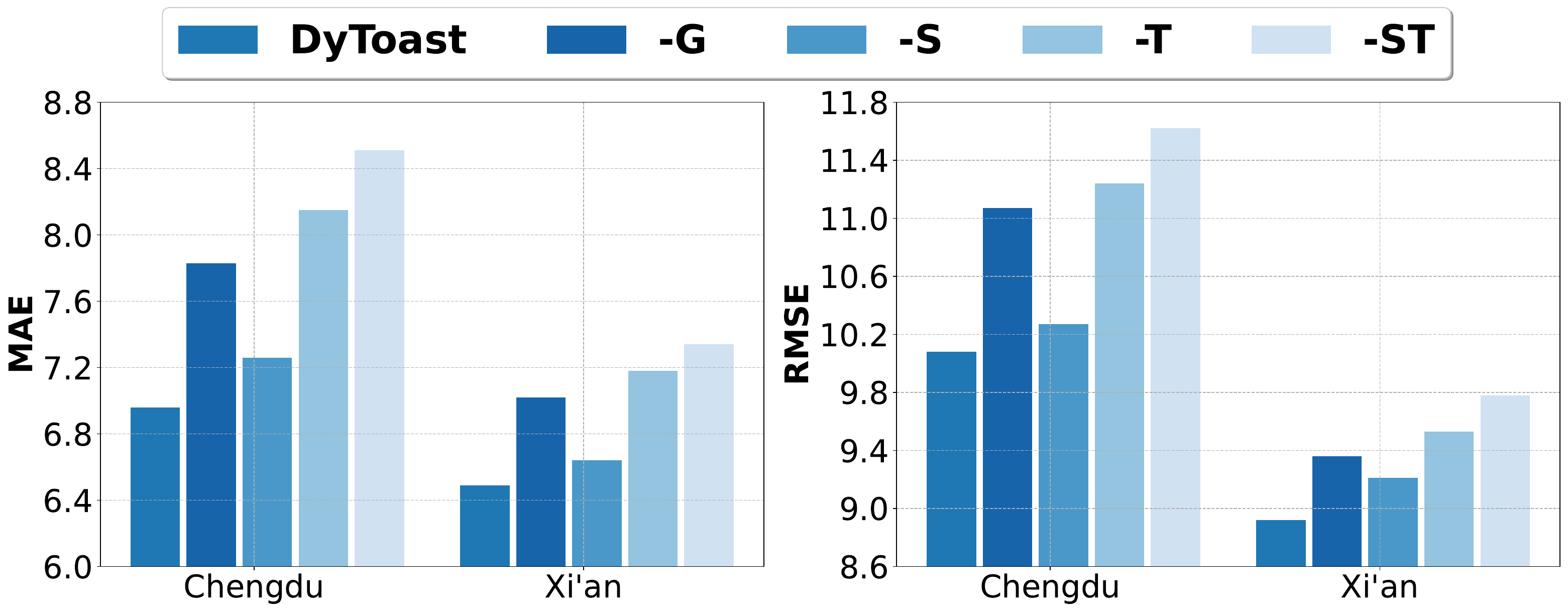}
        \end{minipage}
        \label{subfig:ablation_speed}
    }
    \subfigure[Travel time estimation] {
        \begin{minipage}[t]{1.0\linewidth}
        \centering
        \includegraphics[width=0.95\linewidth]{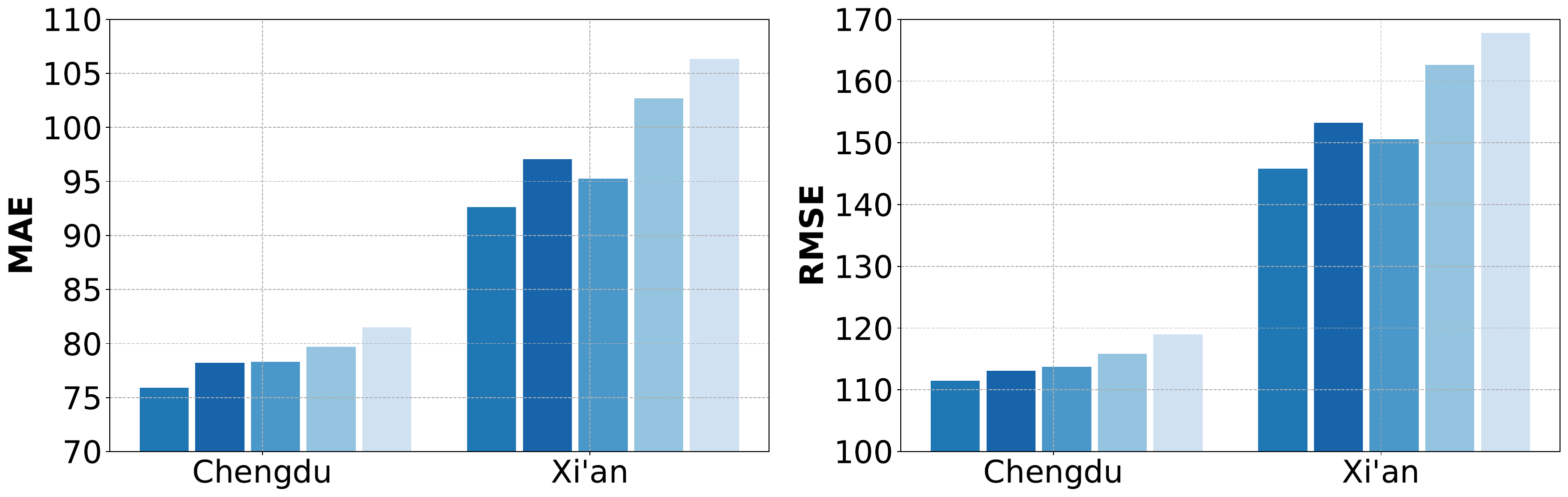}
        \end{minipage}
        \label{subfig:ablation_time}
        \vspace{-2mm}
    }
\vspace{-4mm}
\caption{Ablation study of four model variants without modules for encoding temporal dynamics.}
\label{fig:ablation}
\end{figure}

The results for these model variants on the tasks of traffic speed inference and travel time estimation are shown in Fig.~\ref{fig:ablation}. Based on the results, we can observe that excluding different components from the framework leads to a decrease in performance across both road segment-based and trajectory-based tasks. This highlights the importance of integrating temporal dynamics on road networks from various perspectives to enhance the model performance. Notably, the removal of temporal encoding within the pre-trained Transformer module (-T) indicates a more pronounced impact on model performance than modifications to the traffic-enhanced skip-gram module (-G and -S), demonstrating the benefits of capturing fine-grained temporal correlations in trajectory data. Furthermore, the contributions of these components are complementary, as the removal of multiple modules (i.e. -ST) result in the most significant performance degradation. Overall, the ablation study validates the effectiveness of our proposed techniques to enrich the knowledge on temporal dimension into road network representation learning.  


\begin{table}[tbp]
\centering
\caption{Comparison for temporal encoding techniques on traffic speed inference task.}
\vspace{-2mm}
\resizebox{0.85\linewidth}{!}{
\begin{tabular}{c|cc|cc}
\toprule
Dataset     & \multicolumn{2}{c}{Chengdu} & \multicolumn{2}{c}{Xi'an} \\ \midrule
Metric      & MAE          & RMSE         & MAE         & RMSE        \\ \midrule
T-Emb       & 7.69         & 10.82        & 6.97        & 9.33        \\
Road-Emb    & 7.51         & 10.44        & 6.72        & 9.16        \\
T-Attention & 7.43         & 10.36        & 6.69        & 9.17        \\
CPE         & 7.90         & 11.02        & 7.03        & 9.44        \\ \midrule
DyToast     & \textbf{6.96}         & \textbf{10.08}        & \textbf{6.49}        & \textbf{8.92} \\  \bottomrule
\end{tabular}}
\label{tab:time_encoding}
\end{table}

\subsubsection{Temporal Encoding Techniques}
We further examine the effectiveness of our temporal encoding techniques in \textsf{DyToast} compared to other temporal encoding methods utilized for trajectory data. Specifically, we replace the techniques as described in Section~\ref{subsec:temporal} with four alternative techniques for incorporating time information, while keeping all other components consistent. The details of these techniques are listed as follows:
\begin{itemize}
    \item \textbf{T-Emb}:  it partitions the time into discrete 1-hour intervals and represents each interval with a distinct embedding. These embeddings are applied to all road segments, and subsequently utilized as inputs to the traffic-enhanced skip-gram module. 
    \item \textbf{Road-Emb}: unlike T-Emb, it assigns discrete embeddings based on 1-hour time intervals to each road segment independently. In other words, every road segment possesses its own set of time embeddings.
    \item \textbf{T-Attention}~\cite{ICDE23_traj}: it employs a self-attention mechanism sensitive to time intervals by transforming these intervals into bias terms in attention score calculation in Transformer.
    \item \textbf{CPE}~\cite{CIKM22_traj}: it converts time intervals into kernels, which are applied within convolution operations to incorporate fine-grained temporal information. The results after convolution are utilized as inputs in Transformer.
\end{itemize}

The results of the traffic speed inference task against these compared methods are shown in Table~\ref{tab:time_encoding}, and similar results can  be found on other tasks. From the results, we can make the following observations. First, CPE shows the worst performance, attributed to the misalignment between its original application context in GPS trajectories and the map-matched trajectories in our scenario. Second, despite the simplicity of discrete time embeddings (T-Emb and Road-Emb), they achieve performance improvements for the time-sensitive task. This indicates the benefits of integrating temporal information. While Road-Emb outperforms T-Emb, it also significantly increases the parameter numbers compared to all other methods. Third, T-Attention generally produces the best results among baselines, validating the  effectiveness of enhancing the self-attention mechanism with additional temporal knowledge in trajectory data. Last, \textsf{DyToast} further advances the self-attention mechanism by incorporating sinusoidal function, exhibiting superior properties such as  translation invariance, strong inductive bias and the capacity for continuous temporal modeling. These features facilitate a better capture of temporal dynamics, thereby surpassing these baseline methods in performance.

\subsubsection{Impact of Hyperparameters}

\begin{figure}[t]
  \centering
  \subfigure[Embedding size] {
        \begin{minipage}[t]{1.0\linewidth}
        \centering
        \includegraphics[width=0.95\linewidth]{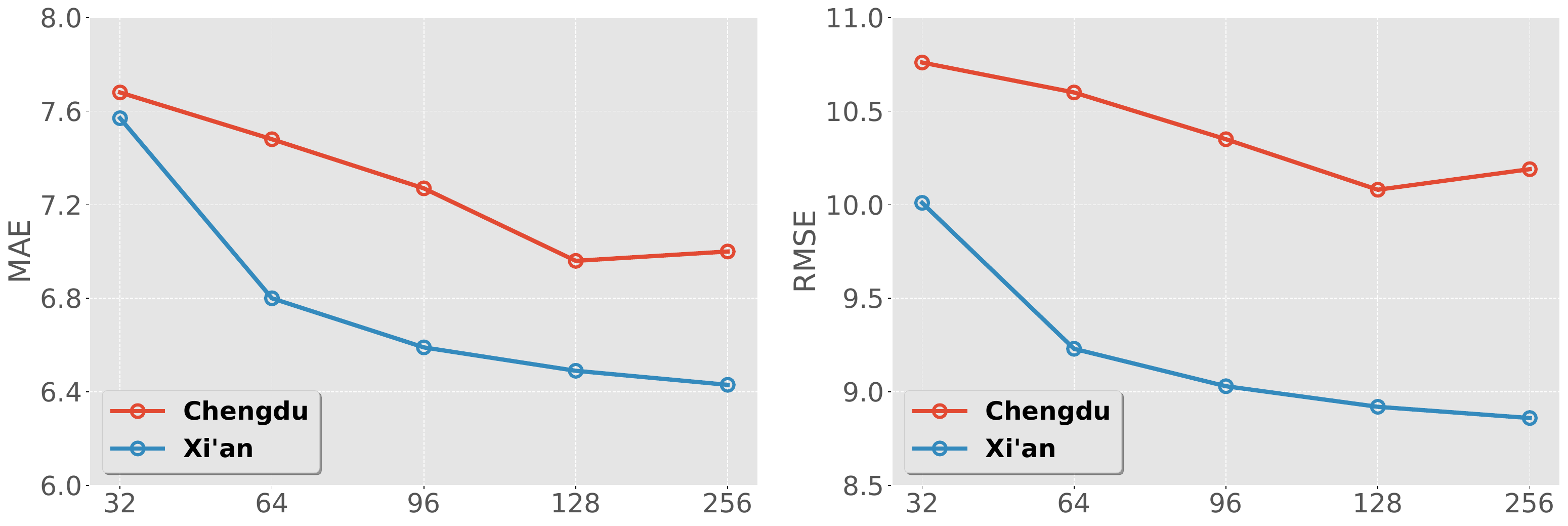}
        \end{minipage}
        \label{subfig:hyper_embedsize}
    }
    \subfigure[Mask ratio] {
        \begin{minipage}[t]{1.0\linewidth}
        \centering
        \includegraphics[width=0.95\linewidth]{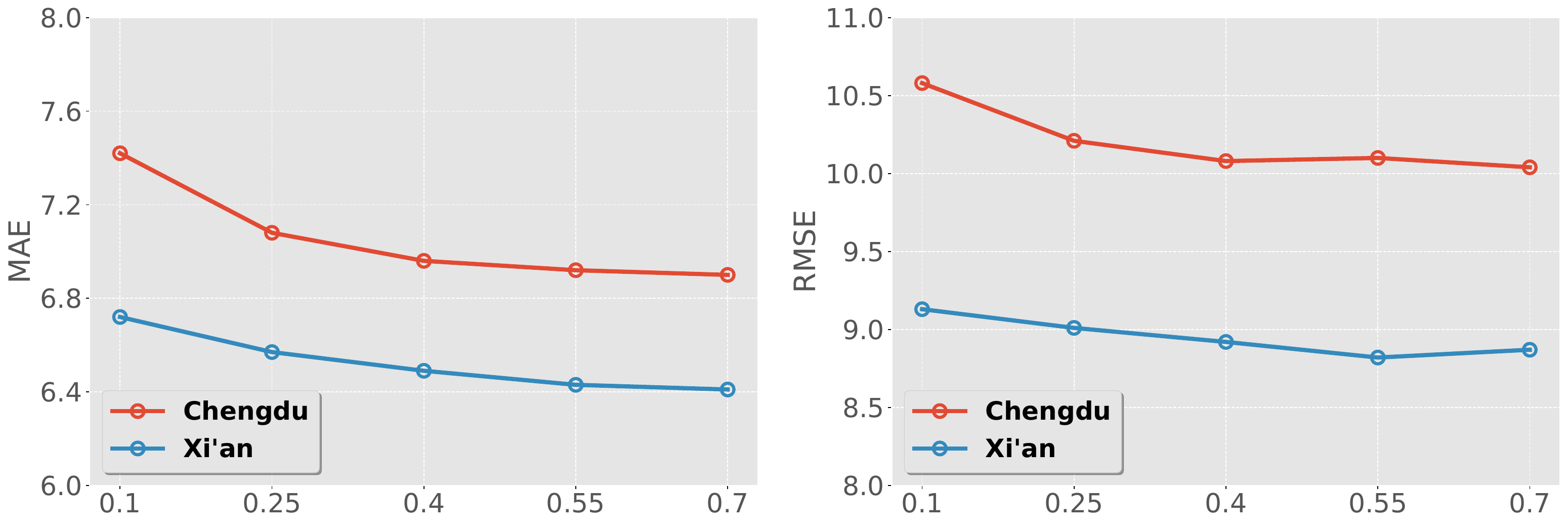}
        \end{minipage}
        \label{subfig:hyper_maskratio}
    }
        \subfigure[Auxiliary loss weight] {
        \begin{minipage}[t]{1.0\linewidth}
        \centering
        \includegraphics[width=0.95\linewidth]{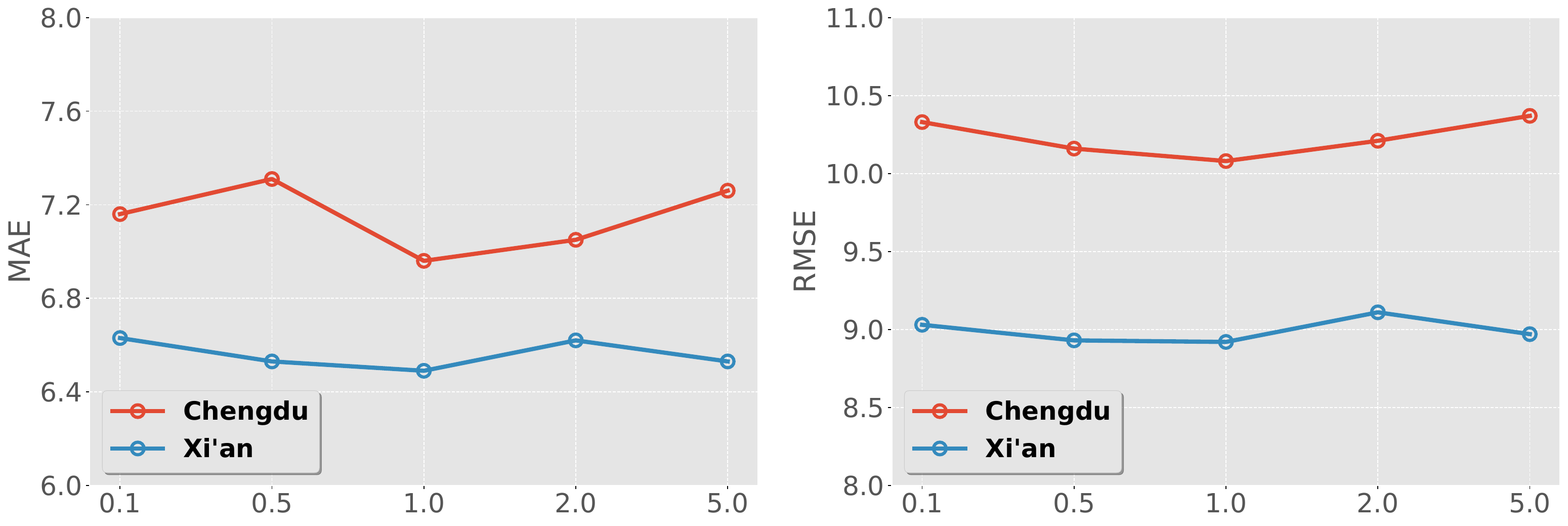}
        \end{minipage}
        \label{subfig:hyper_weight}
    }
    
\caption{Impact of hyperparamters.}
\label{fig:hyper}
\end{figure}

We study the impacts of various hyper-parameters on the model performance, including  embedding size,  mask ratio in Transformer pre-training, and the weight of auxiliary traffic context prediction objective. The results for the traffic speed inference task are presented in Fig.~\ref{fig:hyper}. We omit the results for other tasks since the patterns are found to be similar.

\textbf{Impact of embedding size}. As illustrated in Fig.~\ref{subfig:hyper_embedsize}, an increase in embedding size generally leads to improved model performance. However, when the embedding size exceeds 128, the improvement becomes negligible, or even degrade on the Chengdu dataset probably due to over-fitting issues. Consequently, an embedding size of 128 is set as the default value.

\textbf{Impact of mask ratio}. As illustrated in Fig.~\ref{subfig:hyper_maskratio}, increasing the mask ratio during the Transformer pre-training with trajectory data improves the performance. However, the benefits of increasing the mask ratio becomes saturated after the mask ratio of 0.4, accompanied by more computational cost. Therefore, a mask ratio of 0.4 is chosen as the default setting.

\textbf{Impact of auxiliary loss weight}. As illustrated in Fig.~\ref{subfig:hyper_weight}, selecting either excessively low or high weights for the traffic context prediction objective exhibits negative effect on performance. An optimal value is achieved at a weight of 1.0, which is adopted in our experiments.

\section{Conclusion}
In this paper, we propose a novel framework, \textsf{Toast}, along with its advanced version \textsf{DyToast}, designed to enhance the integration of temporal dynamics for effective road network representation learning. The methods are designed 
to learn generic representations of both road segments trajectories, supporting a wide range of downstream applications, particularly those sensitive to temporal variations. Specifically, our framwork is featured with two modules: a traffic-enhanced skip-gram module to incorporate traffic contexts into the learning process, and a trajectory-enhanced Transformer module to extract the travelling semantics encoded in trajectory data. These modules are further augmented by a unified approach based on trigonometric functions, enabling the capture of temporal dynamics from both time-dependent transportation graphs and trajectory data with fine-grained time interval knowledge. Our experiments demonstrate that the proposed framework consistently outperforms the state-of-the-art road network representation methods on three different tasks within dynamic settings. 

\balance
\bibliographystyle{IEEEtran}
\bibliography{ref}

\begin{thebibliography}{10}
\providecommand{\url}[1]{#1}
\csname url@samestyle\endcsname
\providecommand{\newblock}{\relax}
\providecommand{\bibinfo}[2]{#2}
\providecommand{\BIBentrySTDinterwordspacing}{\spaceskip=0pt\relax}
\providecommand{\BIBentryALTinterwordstretchfactor}{4}
\providecommand{\BIBentryALTinterwordspacing}{\spaceskip=\fontdimen2\font plus
\BIBentryALTinterwordstretchfactor\fontdimen3\font minus \fontdimen4\font\relax}
\providecommand{\BIBforeignlanguage}[2]{{%
\expandafter\ifx\csname l@#1\endcsname\relax
\typeout{** WARNING: IEEEtran.bst: No hyphenation pattern has been}%
\typeout{** loaded for the language `#1'. Using the pattern for}%
\typeout{** the default language instead.}%
\else
\language=\csname l@#1\endcsname
\fi
#2}}
\providecommand{\BIBdecl}{\relax}
\BIBdecl

\bibitem{ICDE_route}
X.~Li, G.~Cong, and Y.~Cheng, ``Spatial transition learning on road networks with deep probabilistic models,'' in \emph{ICDE}, 2020, pp. 349--360.

\bibitem{TKDE22_route}
J.~Wang, N.~Wu, and W.~X. Zhao, ``Personalized route recommendation with neural network enhanced search algorithm,'' \emph{{IEEE} Trans. Knowl. Data Eng.}, vol.~34, no.~12, pp. 5910--5924, 2022.

\bibitem{TKDE_traffic}
S.~Guo, Y.~Lin, H.~Wan, X.~Li, and G.~Cong, ``Learning dynamics and heterogeneity of spatial-temporal graph data for traffic forecasting,'' \emph{IEEE Transactions on Knowledge and Data Engineering}, 2021.

\bibitem{ICDE19_trafficinfer}
J.~Hu, C.~Guo, B.~Yang, and C.~S. Jensen, ``Stochastic weight completion for road networks using graph convolutional networks,'' in \emph{ICDE}, 2019, pp. 1274--1285.

\bibitem{WWW22_homophily}
L.~Du, X.~Shi, Q.~Fu, X.~Ma, H.~Liu, S.~Han, and D.~Zhang, ``{GBK-GNN:} gated bi-kernel graph neural networks for modeling both homophily and heterophily,'' in \emph{The {ACM} Web Conference 2022}, 2022, pp. 1550--1558.

\bibitem{CIKM23_homophily2}
M.~Gu, G.~Yang, S.~Zhou, N.~Ma, J.~Chen, Q.~Tan, M.~Liu, and J.~Bu, ``Homophily-enhanced structure learning for graph clustering,'' in \emph{{CIKM}}, 2023, pp. 577--586.

\bibitem{Deepwalk}
B.~Perozzi, R.~Al{-}Rfou, and S.~Skiena, ``Deepwalk: online learning of social representations,'' in \emph{KDD}, 2014, pp. 701--710.

\bibitem{node2vec}
A.~Grover and J.~Leskovec, ``node2vec: Scalable feature learning for networks,'' in \emph{KDD}, 2016, pp. 855--864.

\bibitem{GNN_Survey}
Z.~Wu, S.~Pan, F.~Chen, G.~Long, C.~Zhang, and P.~S. Yu, ``A comprehensive survey on graph neural networks,'' \emph{IEEE Transactions on Neural Networks and Learning Systems}, 2020.

\bibitem{GNN_oversmooth}
Q.~Li, Z.~Han, and X.~Wu, ``Deeper insights into graph convolutional networks for semi-supervised learning,'' in \emph{AAAI}, 2018, pp. 3538--3545.

\bibitem{Bigdata18_roadrep}
T.~S. Jepsen, C.~S. Jensen, T.~D. Nielsen, and K.~Torp, ``On network embedding for machine learning on road networks: {A} case study on the danish road network,'' in \emph{{IEEE} BigData 2018,}, 2018, pp. 3422--3431.

\bibitem{SIGSPATIAL19_roadrep}
M.~Wang, W.~Lee, T.~Fu, and G.~Yu, ``Learning embeddings of intersections on road networks,'' in \emph{SIGSPATIAL}, 2019, pp. 309--318.

\bibitem{TIST21}
M.~Wang, W.~C. Lee, T.~Fu, and G.~Yu, ``On representation learning for road networks,'' \emph{{ACM} Trans. Intell. Syst. Technol.}, vol.~12, no.~1, pp. 11:1--11:27, 2021.

\bibitem{SIGSPATIAL_roadrep}
T.~S. Jepsen, C.~S. Jensen, and T.~D. Nielsen, ``Graph convolutional networks for road networks,'' in \emph{SIGSPATIAL}, 2019, pp. 460--463.

\bibitem{RoadGNN_KDD20}
N.~Wu, W.~X. Zhao, J.~Wang, and D.~Pan, ``Learning effective road network representation with hierarchical graph neural networks,'' in \emph{KDD}, 2020, pp. 6--14.

\bibitem{TITS22_roadrep}
T.~S. Jepsen, C.~S. Jensen, and T.~D. Nielsen, ``Relational fusion networks: Graph convolutional networks for road networks,'' \emph{{IEEE} Trans. Intell. Transp. Syst.}, vol.~23, no.~1, pp. 418--429, 2022.

\bibitem{IJCAI17_traj}
H.~Wu, Z.~Chen, W.~Sun, B.~Zheng, and W.~Wang, ``Modeling trajectories with recurrent neural networks,'' in \emph{IJCAI}, C.~Sierra, Ed., 2017, pp. 3083--3090.

\bibitem{TIST20}
T.~Fu and W.~Lee, ``Trembr: Exploring road networks for trajectory representation learning,'' \emph{{ACM} Trans. Intell. Syst. Technol.}, vol.~11, no.~1, pp. 10:1--10:25, 2020.

\bibitem{BERT}
J.~Devlin, M.~Chang, K.~Lee, and K.~Toutanova, ``{BERT:} pre-training of deep bidirectional transformers for language understanding,'' in \emph{NAACL}, 2019, pp. 4171--4186.

\bibitem{Toast}
Y.~Chen, X.~Li, G.~Cong, Z.~Bao, C.~Long, Y.~Liu, A.~K. Chandran, and R.~Ellison, ``Robust road network representation learning: When traffic patterns meet traveling semantics,'' in \emph{{CIKM}}, 2021, pp. 211--220.

\bibitem{EDBT23_roadrep}
Y.~Chang, E.~Tanin, X.~Cao, and J.~Qi, ``Spatial structure-aware road network embedding via graph contrastive learning,'' in \emph{{EDBT}}, 2023, pp. 144--156.

\bibitem{CIKM22_roadrep}
Z.~Mao, Z.~Li, D.~Li, L.~Bai, and R.~Zhao, ``Jointly contrastive representation learning on road network and trajectory,'' in \emph{{CIKM}}, 2022, pp. 1501--1510.

\bibitem{TKDD23_roadrep}
L.~Zhang and C.~Long, ``Road network representation learning: {A} dual graph-based approach,'' \emph{{ACM} Trans. Knowl. Discov. Data}, vol.~17, no.~9, pp. 121:1--121:25, 2023.

\bibitem{PAKDD23_roadrep}
S.~Schestakov, P.~Heinemeyer, and E.~Demidova, ``Road network representation learning with vehicle trajectories,'' in \emph{{PAKDD}}, 2023, pp. 57--69.

\bibitem{SIGSPATIAL21_attribute}
S.~Wozniak and P.~Szymanski, ``hex2vec: Context-aware embedding {H3} hexagons with openstreetmap tags,'' in \emph{{SIGSPATIAL}}, 2021, pp. 61--71.

\bibitem{VLDB22_traveltime}
H.~Yuan, G.~Li, and Z.~Bao, ``Route travel time estimation on {A} road network revisited: Heterogeneity, proximity, periodicity and dynamicity,'' \emph{Proc. {VLDB} Endow.}, vol.~16, no.~3, pp. 393--405, 2022.

\bibitem{traj_survey}
S.~Wang, Z.~Bao, J.~S. Culpepper, and G.~Cong, ``A survey on trajectory data management, analytics, and learning,'' \emph{{ACM} Comput. Surv.}, vol.~54, no.~2, pp. 39:1--39:36, 2021.

\bibitem{ICDE20_trajanomaly}
Y.~Liu, K.~Zhao, G.~Cong, and Z.~Bao, ``Online anomalous trajectory detection with deep generative sequence modeling,'' in \emph{ICDE}, 2020, pp. 949--960.

\bibitem{TKDE20_destination}
Z.~Yang, H.~Sun, J.~Huang, Z.~Sun, H.~Xiong, S.~Qiao, Z.~Guan, and X.~Jia, ``An efficient destination prediction approach based on future trajectory prediction and transition matrix optimization,'' \emph{{IEEE} Trans. Knowl. Data Eng.}, vol.~32, no.~2, pp. 203--217, 2020.

\bibitem{KDD21_trajflow}
B.~Hui, D.~Yan, H.~Chen, and W.~Ku, ``Trajnet: {A} trajectory-based deep learning model for traffic prediction,'' in \emph{{KDD}}, F.~Zhu, B.~C. Ooi, and C.~Miao, Eds., 2021, pp. 716--724.

\bibitem{AAAI21_trajflow}
M.~Li, P.~Tong, M.~Li, Z.~Jin, J.~Huang, and X.~Hua, ``Traffic flow prediction with vehicle trajectories,'' in \emph{{AAAI}}, 2021, pp. 294--302.

\bibitem{IJCAI22_region}
S.~Wu, X.~Yan, X.~Fan, S.~Pan, S.~Zhu, C.~Zheng, M.~Cheng, and C.~Wang, ``Multi-graph fusion networks for urban region embedding,'' in \emph{{IJCAI}}, 2022, pp. 2312--2318.

\bibitem{TKDE23_regioon}
L.~Zhang, C.~Long, and G.~Cong, ``Region embedding with intra and inter-view contrastive learning,'' \emph{{IEEE} Trans. Knowl. Data Eng.}, vol.~35, no.~9, pp. 9031--9036, 2023.

\bibitem{MapMatching}
C.~Yang and G.~Gidofalvi, ``Fast map matching, an algorithm integrating hidden markov model with precomputation,'' \emph{International Journal of Geographical Information Science}, vol.~32, no.~3, pp. 547--570, 2018.

\bibitem{word2vec}
T.~Mikolov, I.~Sutskever, K.~Chen, G.~S. Corrado, and J.~Dean, ``Distributed representations of words and phrases and their compositionality,'' in \emph{NIPS}, 2013, pp. 3111--3119.

\bibitem{NIPS17_Attn}
A.~Vaswani, N.~Shazeer, N.~Parmar, J.~Uszkoreit, L.~Jones, A.~N. Gomez, L.~Kaiser, and I.~Polosukhin, ``Attention is all you need,'' in \emph{NIPS}, 2017, pp. 5998--6008.

\bibitem{struct2vec}
L.~F.~R. Ribeiro, P.~H.~P. Saverese, and D.~R. Figueiredo, ``\emph{struc2vec}: Learning node representations from structural identity,'' in \emph{KDD}, 2017, pp. 385--394.

\bibitem{GPT3}
T.~B. Brown, B.~Mann, and et~al., ``Language models are few-shot learners,'' \emph{CoRR}, vol. abs/2005.14165, 2020.

\bibitem{ImageBERT}
J.~Lu, D.~Batra, D.~Parikh, and S.~Lee, ``Vilbert: Pretraining task-agnostic visiolinguistic representations for vision-and-language tasks,'' in \emph{NeurIPS}, 2019, pp. 13--23.

\bibitem{VideoBERT}
C.~Sun, A.~Myers, C.~Vondrick, K.~Murphy, and C.~Schmid, ``Videobert: {A} joint model for video and language representation learning,'' in \emph{ICCV}, 2019, pp. 7463--7472.

\bibitem{ALBERT}
Z.~Lan, M.~Chen, S.~Goodman, K.~Gimpel, P.~Sharma, and R.~Soricut, ``{ALBERT:} {A} lite {BERT} for self-supervised learning of language representations,'' in \emph{{ICLR}}, 2020.

\bibitem{polynomial}
A.~Pinkus, ``Weierstrass and approximation theory,'' \emph{Journal of Approximation Theory}, vol. 107, no.~1, pp. 1--66, 2000.

\bibitem{word2fun}
B.~Wang, E.~D. Buccio, and M.~Melucci, ``Word2fun: Modelling words as functions for diachronic word representation,'' in \emph{{NeurIPS 2021}}, 2021, pp. 2861--2872.

\bibitem{relative_position}
P.~Shaw, J.~Uszkoreit, and A.~Vaswani, ``Self-attention with relative position representations,'' in \emph{NAACL-HLT}, 2018, pp. 464--468.

\bibitem{NIPS_kernel}
A.~Rahimi and B.~Recht, ``Random features for large-scale kernel machines,'' in \emph{{NIPS}}, 2007, pp. 1177--1184.

\bibitem{OSM}
{OpenStreetMap}, [Online]. \url{https://www. openstreetmap.org/}.

\bibitem{GCN}
T.~N. Kipf and M.~Welling, ``Semi-supervised classification with graph convolutional networks,'' in \emph{ICLR}, 2017.

\bibitem{GAT}
P.~Velickovic, G.~Cucurull, A.~Casanova, A.~Romero, P.~Li{\`{o}}, and Y.~Bengio, ``Graph attention networks,'' in \emph{ICLR}, 2018.

\bibitem{ICDE23_traj}
J.~Jiang, D.~Pan, H.~Ren, X.~Jiang, C.~Li, and J.~Wang, ``Self-supervised trajectory representation learning with temporal regularities and travel semantics,'' in \emph{{ICDE}}, 2023, pp. 843--855.

\bibitem{CIKM22_traj}
Y.~Liang, K.~Ouyang, Y.~Wang, X.~Liu, H.~Chen, J.~Zhang, Y.~Zheng, and R.~Zimmermann, ``Trajformer: Efficient trajectory classification with transformers,'' in \emph{{CIKM}}, 2022, pp. 1229--1237.

\end{thebibliography}


 




\end{document}